\documentclass[11pt,a4paper]{article}
\usepackage[hyperref]{emnlp-ijcnlp-2019}

\usepackage{url}
\usepackage{times}
\usepackage{latexsym}
\usepackage{tabularx}
\usepackage{graphicx}
\usepackage{booktabs}
\usepackage{amsfonts}       
\usepackage{nicefrac}       
\usepackage{microtype}      
\usepackage{multirow}
\usepackage{xcolor}
\usepackage{caption}
\usepackage{amsmath}
\usepackage{float} 
\usepackage{ragged2e}

\usepackage{enumitem}
\mathchardef\mhyphen="2D

\aclfinalcopy 

\newcommand\mf[1]{\mathbf{#1}}
\newcommand\ourmodel{Sparse IB}
\newcommand\sparsenorm{Sparse Norm}
\newcommand\sparsenormthreshold{Sparse Norm-C}
\newcommand\nosparsity{No Sparsity}
\newcommand\fullcontext{Full}
\newcommand\goldcontext{Gold}
\newcommand\beer{BeerAdvocate}

\title{An Information Bottleneck Approach for \\ Controlling Conciseness in Rationale Extraction}

\author{Bhargavi Paranjape$^{\dagger}$ \quad Mandar Joshi$^{\dagger}$ \quad John Thickstun$^{\dagger}$ \\ \bf Hannaneh Hajishirzi$^{\dagger\epsilon}$ \quad \bf Luke Zettlemoyer$^{\dagger}$ \\ $^{\dagger}$ Allen School of Computer Science \& Engineering, University of Washington, Seattle, WA \\  $^{\epsilon}$Allen Institute of Artificial Intelligence, Seattle \\ 
{\tt \{bparan,mandar90,thickstn,hannaneh,lsz\}@cs.washington.edu}}

\date{}

\begin{document}
\maketitle
\begin{abstract}
Decisions of complex models for language understanding can be explained by limiting the inputs they are provided to a relevant subsequence of the original text --- a \emph{rationale}.
Models that condition predictions on a concise rationale, while being more interpretable, tend to be less accurate than models that are able to use the entire context.
In this paper, we show that it is possible to better manage the trade-off 
between concise explanations and high task accuracy by optimizing a bound on the Information Bottleneck (IB) objective.
Our approach jointly learns an explainer that predicts sparse binary masks over input sentences without explicit supervision, and an end-task predictor that considers only the residual sentences.
Using IB, we derive a learning objective that allows direct control of mask sparsity levels through a tunable sparse prior.
Experiments on the ERASER benchmark demonstrate significant gains over previous work  
for both task performance and agreement with human rationales.
Furthermore, we find that in the semi-supervised setting, a modest amount of gold rationales ($25\%$ of training examples with gold masks) can close the performance gap with a model that uses the full input.\footnote{Our code is available at \url{https://github.com/bhargaviparanjape/explainable_qa}}
\end{abstract}

\section{Introduction}
\begin{figure}[ht]
     \centering
     \includegraphics[scale=0.42]{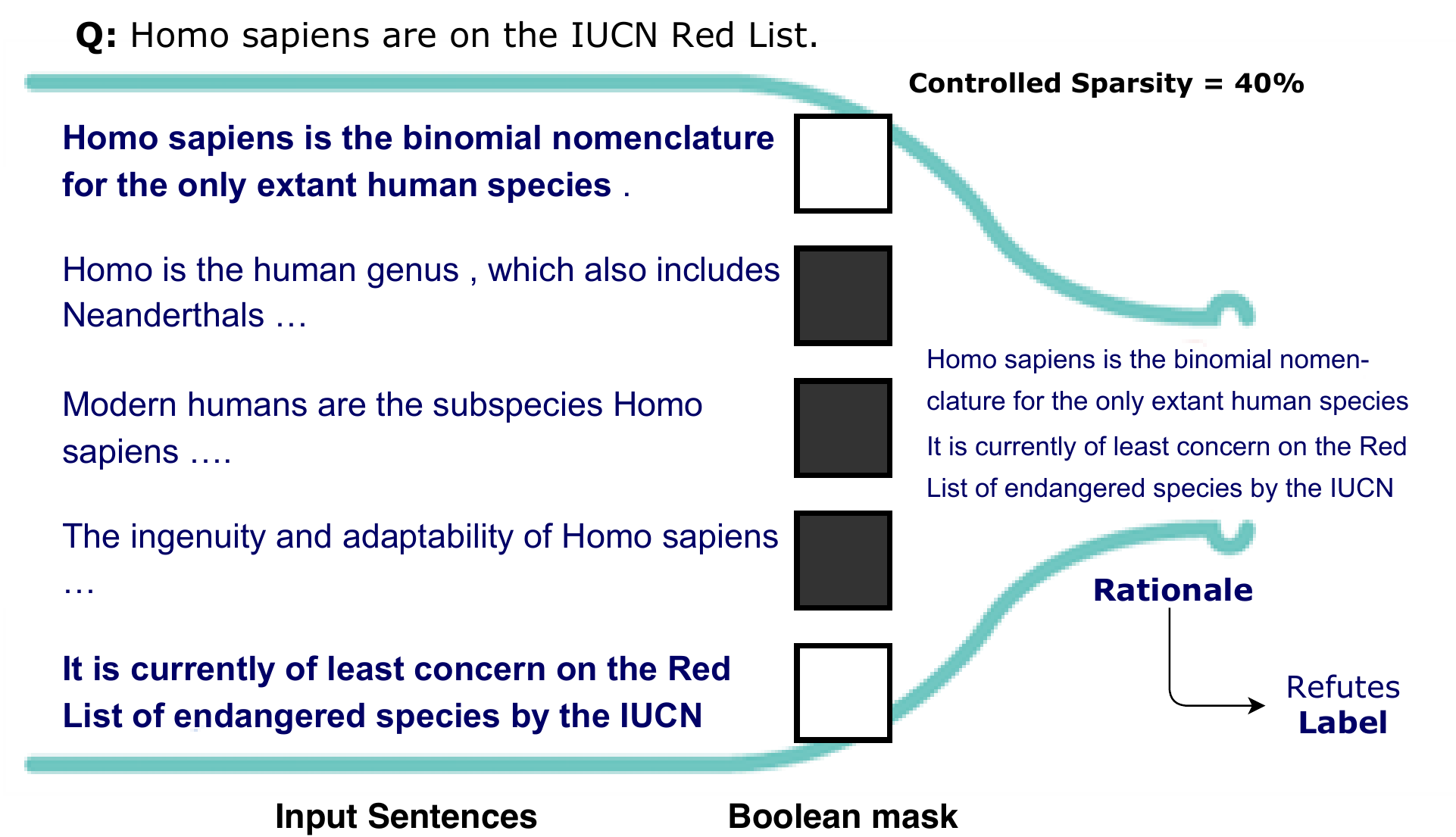}
     \caption{Our Information Bottleneck-based approach extracts concise rationales that are minimally informative about the original input, and  maximally informative about the label through fine-grained control of sparsity in the bottleneck ($0.4$ in this fact verification example). End-task prediction is conditioned only on the bottlenecked input.
     }
     \label{fig:Fig1}
 \end{figure}

A rationale is a short yet sufficient part of the input text that can explain model decisions for a range of language understanding tasks~\cite{lei2016rationalizing}. Models can be \emph{faithful} to a rationale by only using the selected text as input for end-task prediction~\cite{deyoung2019eraser}.
However, there is almost always a trade-off between interpretable models that learn to extract \emph{sparse} rationales and more \emph{accurate} models that are able to use the full context but provide little explanation for their predictions ~\cite{lei2016rationalizing,  weld2019challenge}.
In this paper, we show that it is possible to better manage this trade-off by optimizing a novel bound on the Information Bottleneck \cite{tishby99information} objective  (Figure~\ref{fig:Fig1}).

We follow recent work in representing rationales as binary masks over the input text~\cite{lei2016rationalizing,bastings2019interpretable}. During learning, it is common to encourage sparsity by minimizing a norm on the rationale masks (e.g. $L_0$ or $L_1$)~\cite{lei2016rationalizing, bastings2019interpretable}. 
It is often challenging to control the sparsity-accuracy trade-off in norm-minimization methods; we show that these methods seem to push too directly for sparsity at the expense of accuracy (Section~\ref{subsec:analysis}).
Our approach, in contrast, allows more control through a prior that specifies task-specific target sparsity levels that should be met in expectation across the training set. 

 More specifically, we formalize the problem of inducing controlled sparsity in the mask using the Information Bottleneck (IB) principle. 
 Our approach seeks to extract a rationale as an 
 optimal \emph{compressed} intermediate representation (the bottleneck) that is both (1) minimally informative about the original input, and (2) maximally informative about the output class.  We derive a novel variational bound on the IB objective for our case where we constrain the intermediate representation to be a concise subsequence of the input, thus ensuring its interpretablity.


Our model consists of an \emph{explainer} that extracts a rationale from the input, and an \emph{end-task predictor} that predicts the output based only on the extracted rationale. 
Our IB-based training objective guarantees sparsity by minimizing the Kullback–Leibler (KL) divergence between the explainer mask probability distribution and a prior distribution with controllable sparsity levels.
This prior probability affords us tunable fine-grained control over sparsity, and allows us to bias the proportion of the input to be used as rationale. We show that, unlike norm-minimization methods, our KL-divergence objective is able to consistently extract rationales with the specified sparsity levels.

Across five tasks from the ERASER interpretability benchmark \cite{deyoung2019eraser} and the \beer~dataset \cite{mcauley2012learning}, 
our IB-based sparse prior objective has significant gains over previous norm-minimization techniques ---
up to 5\% relative improvement in task performance metrics and 6\% to 80\% relative improvement in agreement with human rationale annotations. 
Our interpretable model achieves task performance within $10\%$ of a model of comparable size that uses the entire input. Furthermore, we find that in the semi-supervised setting, adding a small proportion of gold rationale annotations (approximately $25\%$ of the training examples) bridges this gap --- we are able to build an interpretable  
model without compromising performance.
\section{Method}
\subsection{Task and Method Overview}
\label{sec:prelims}
We assume supervised text classification or regression data that contains tuples of the form $(x,y)$. The input document $x$ can be decomposed into a sequence of sentences $x = (x_{1}, x_{2}, \ldots , x_{n})$ and $y$ is the category, answer choice, or target value to predict.
Our goal is to learn a model that not only predicts $y$, but also extracts a rationale or explanation $z$---a latent  \emph{subsequence} of sentences in $x$ with the following properties:
\begin{enumerate}[noitemsep,topsep=2pt,parsep=0pt,partopsep=2pt]
    \item Model prediction $y$ should rely entirely on $z$ and not on its complement $x \backslash z$ --- \emph{faithfulness} \cite{deyoung2019eraser}.
    \item $z$ must be concise, i.e., it should contain as few sentences as possible without sacrificing the ability to correctly predict $y$.
\end{enumerate}

Following \citet{lei2016rationalizing}, our interpretable model learns a boolean mask $ m = (m_{1}, m_{2}, \ldots , m_{n})$ over the sentences in $x$, where $m_j \in \{0,1\}$ is a discrete binary variable. To enforce (1), the masked input $z = m \odot x = (m_{1} \cdot x_{1},m_{2} \cdot x_{2}, \ldots , m_{n} \cdot x_{n})$ is used to predict $y$. Conciseness is attained using an information bottleneck.

\subsection{Formalizing Interpretability Using Information Bottleneck}
\label{sec:formalizing}


\paragraph{Background}
The Information Bottleneck (IB) method is used to learn an optimal compression model that transmits information from a random variable $X$ to another random variable $Y$ through a compressed representation $Z$. The IB objective is to minimize the following:
\begin{equation}
    L_{IB} = I(X, Z) - \beta I(Z, Y),
\label{ib}
\end{equation}
where $I(\cdot,\cdot)$ is mutual information. This objective encourages $Z$ to only retain as much information about $X$ as is needed to predict $Y$. The hyperparameter $\beta$ controls the trade-off between retaining information about either $X$ or $Y$ in $Z$.
\citet{alemi2016deep} derive the following variational bound on Equation \ref{ib}:\footnote{For brevity and clarity, objectives are shown for a single data point. More details of this bound can be found in Appendix \ref{ib_alemi} and \citet{alemi2016deep}.}
\begin{multline}
        L_{VIB} = \underbrace{\mathbb{E}_{z \sim p_{\theta}(z|x)}[-\log{q_{\phi}(y|z)}]}_\text{Task Loss} + \\
    \underbrace{\beta KL[p_{\theta}(z|x), r(z)],}_\text{Information Loss}
\label{alemi_ib}
\end{multline}

\noindent where $q_{\phi}(y|z)$ is a  parametric approximation to the true likelihood $p(y|z)$; 
$r(z)$, the prior probability of $z$, approximates the marginal $p(z)$; and $p_{\theta}(z|x)$ is the parametric posterior distribution over $z$.

The information loss term in Equation~\ref{alemi_ib} reduces $I(X, Z)$ by decreasing the KL divergence\footnote{\label{zisdense}To analytically compute the KL-divergence term, the posterior and prior distributions over $z$ are typically  K-dimensional multivariate normal distributions. Compression is achieved by setting $K << D$, the input dimension of $X$.} between the posterior distribution $p_{\theta}(z|x)$ that depends on $x$ and a prior distribution $r(z)$ that is independent of $x$. The task loss encourages predicting the correct label $y$ from $z$ to increase $I(Z,Y)$. 

\paragraph{Our Variational Bound for Interpretability}
The learned bottleneck representation $z$, found via Equation~\ref{alemi_ib}, is not human-interpretable as $z$ is typically a compressed continuous vector representation of input $x$.\footnotemark[3]
To ensure interpretability of $z$, we define the interpretable latent representation as $z := m \odot x$, where $m$ is a boolean mask on the input sentences in $x$.
We assume that the mask variables $m_j$ over individual sentences are conditionally independent given the input $x$, i.e. the posterior $p_\theta(m|x) = \prod_j p_\theta(m_j|x)$, where $p_\theta(m_j|x) = \text{Bernoulli}(\theta_j(x))$ and $j$ indexes sentences in the input text.\footnote{We use Bernoulli distribution for $m_j$ in this work, but any binary distribution for which KL divergence can be analytically computed can be used.}
Because $z := m \odot x$, the posterior distribution over $z$ is a mixture of dirac-delta distributions:
\[
p_\theta(z_j|x) = (1-\theta_j(x))\delta(z_j) + \theta_j(x)\delta(z_j - x_j),
\]
where $\delta(x-c)$ is the dirac-delta probability distribution that is zero everywhere except at $c$.

For the prior, we assume a fixed Bernoulli distribution over mask variables. For instance, $r(m_j) = \text{Bernoulli}(\pi)$ for some constant $\pi \in (0,1)$. This also induces a fixed distribution on $z$ via the definition $z := m \odot x$. 
Instead of using an expressive $r(z)$ to approximate $p(z)$, we use a non-parametric prior $r(z)$ to force the marginal $p(z)$ of the learned distribution over $z$ to approximately equal $\pi$. 
Our characterization of the prior and the posterior achieves compression of the input via \emph{sparsity} in the latent representation, in contrast to compression via dimensionality reduction \cite{alemi2016deep}.

For the intermediate representation $z := m \odot x$, we can decompose $\text{KL}(p_\theta(z_j|x),r(z_j))$ as:
\begin{align*}
\text{KL}(p_\theta(m_j|x),r(m_j)) + \pi H(x)
\end{align*}
Since the entropy of the input, $\pi H(x)$, is a constant with respect to $\theta$, it can be dropped.
Hence, we obtain the following variational bound on IB with interpretability constraints over $z$, derived in more detail in Appendix~\ref{iib_spike_spike}:
\begin{multline}
L_{IVIB} = \mathbf{E}_{m \sim p_{\theta}(m|x)}[-\log{q_{\phi}(y|m \odot x)}] + \\
\beta \sum_j KL[p_{\theta}(m_j|x)||r(m_j)]
\label{ivib}
\end{multline}
The first term is the expected cross-entropy term for the task which can be computed by drawing samples $m \sim p_{\theta}(m|x)$. The second information-loss term encourages the mask $m$ to be independent of $x$ by reducing the KL divergence of its posterior $p_{\theta}(m|x)$ from a prior $r(m)$ that is independent of $x$.  However, this does not necessarily remove information about $x$ in $z = x \odot m$. For instance, a mask consisting of all ones is independent of $x$, but in this case $z = x$ and the rationale is no longer concise. In the following section, we present a simple way to avoid this degenerate case in practice by appropriately fixing the value of $\pi$.

\subsection{The Sparse Prior Objective}
\label{lsp_text}
The key to ensuring that $z = m \odot x$ is strictly a subsequence of $x$ lies in the fact that $r(m_j) = \pi$ is our prior belief about the probability of a sentence being important for prediction. For instance, if humans annotate $10\%$ of the input text as a rationale, we can fix our prior belief that a sentence should be a part of the mask as $r(m_j) = \pi = 0.1 \; \forall j$.
IB allows us to \emph{control} the amount of sparsity in the mask that is eventually sampled from the learned distribution $p_\theta(m|x)$ in several ways. $\pi$ can be estimated as the expected sparsity of the mask from expert rationale annotations. If such a statistic is not available, it can be explicitly tuned for the desired trade-off between end task performance and rationale length. In this work, we assume $\pi \in (0,0.5)$ so that the sampled mask is sparse. We refer to this training objective with tunable $r(m) = \pi$ as the sparse prior (\ourmodel) method in our experiments. In Appendix~\ref{kuma}, we also discuss explicitly learning the value of $\pi$.  


\begin{figure}
    \centering
    \includegraphics[scale=0.38]{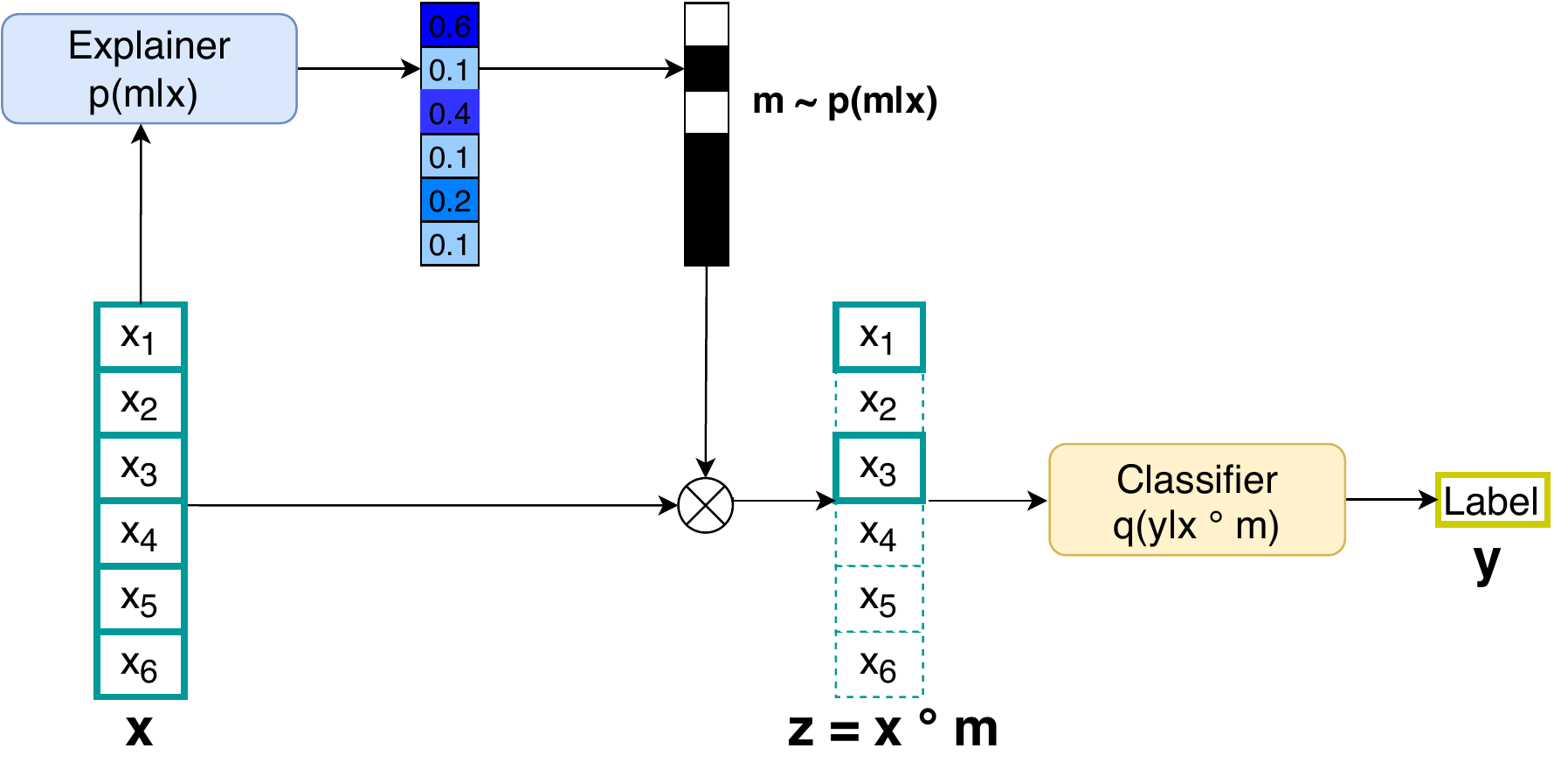}
    \caption{Architecture: The explainer extracts a rationale from the input using a binary mask, and an end-task predictor predicts the output based only on the extracted rationale.}
    \label{fig:model}
\end{figure}
\section{Model}

To optimize for objective \ref{ivib}, the posterior distribution estimator $p_{\theta}()$ and label likelihood estimator $q_{\phi}()$ are instantiated as the \emph{explainer} and \emph{end-task predictor} neural models respectively. Two different pre-trained transformers~\cite{devlin2019bert} are used to initialize both models.

\subsection{Architecture}
\label{sec:arch}

\paragraph{Explainer $p_{\theta}(z|x)$:} Given an input $x = x_1, x_2, \ldots, x_n$ consisting of $n$ sentences, the explainer produces a binary mask $m \in \{0, 1\}^n$ over the input sentences which is used to derive a rationale $z = m \odot x$. It maps every sentence $x_j$ to its probability, $p_{\theta}(m_j|x)$ of being selected as part of $z$ where $p(\cdot)$ is a binary distribution. The explainer contextualizes the input sequence $x$ at the token level, and produces sentence representations $\mf{x} = ( \mf{x_1}, \mf{x_2}, \ldots , \mf{x_n})$ where $\mf{x_j}$ is obtained by concatenating the contextualized representations of the first and last tokens in sentence $x_j$. A linear layer is used to transform these representations into logits (log probabilities) of a Bernoulli distribution. We choose the Bernoulli distribution since its sample can be reparameterized as described in Section \ref{method:reparameterize}, and we can analytically compute the KL-divergence term between two Bernoulli distributions. In Appendix \ref{kuma}, we also experiment with the Kumaraswamy distribution \cite{fletcher1996estimation} used in \cite{bastings2019interpretable}.
The mask $m \in \{0,1\}^n$  is constructed by independently sampling each $m_j$ from $p(m_j|x)$. 

\paragraph{End-task Predictor $q_{\phi}(y|z)$:} We define $z$ as the rationale representation $z = m \odot \mf{x}$, an element-wise dot product between $m_j$ and the corresponding sentence representation $\mf{x_j}$. The end-task predictor uses $z$ to predict the output variable $y$. The same hard attention mask $m$ is applied to all end-task transformer layers at every head to ensure prediction relies only on $m \odot x$. The predictor further consists of a log-linear classifier layer over the $\texttt{[CLS]}$ token, similar to \citet{devlin2019bert}.
When an optional query sequence is available for datasets like BoolQ, we do not mask it as it is assumed to be essential to predict $y$ (see Appendix~\ref{appendix:modeling} for implementation details). 

\subsection{Training and Inference}
\label{method:reparameterize}
The sampling operation of the discrete binary variable $m_j \in \{0,1\}$ in Section \ref{sec:arch} is not differentiable. \citet{lei2016rationalizing} use a simple Bernoulli distribution with REINFORCE \cite{williams1992simple} to overcome non-differentiability.  We found REINFORCE to be quite unstable with high variance in results. Instead, we employ reparameterization \cite{kingma2015variational} to facilitate end-to-end differentiability of our approach.
 We use the Gumbel-Softmax reparameterization \cite{jang2017categorical} for categorical (here, binary) distributions to reparameterize the Bernoulli variables $m_j$.
The reparameterized binary variable $m_j^*$ is generated as follows:
$$m_j^* = \sigma \left(\frac{\log{p(m_j|x)} + g_j}{\tau}\right),$$
where $\sigma$ is the Sigmoid function, $\tau$ is a hyperparameter for the temperature of the Gumbel-Softmax function, and $g_j$ is a random sample from the Gumbel(0,1) distribution \cite{gumbel1948statistical}. $m_j^* \in (0,1)$ is a continuous and differentiable approximation to $m_j$ with low variance.

During inference, we extract the top $\pi\%$ sentences with largest $p_{\theta}(m_j|x)$ values, where $\pi$ corresponds to the threshold hyperparameter described in Section \ref{lsp_text}. 
Previous work \cite{lei2016rationalizing,bastings2019interpretable} samples from $p(m|x)$ during inference. Such an inference strategy is non-deterministic, making comparison of different masking strategies difficult. Moreover, it is possible to appropriately scale $p(m_j|x)$ values to obtain better inference results, thereby not reflecting if $p(m_j|x) \; \forall j$ are correctly ordered.
By allowing a fixed budget of $\pi\%$ per example, we are able to fairly compare approaches in Section \ref{baselines}. 

\subsection{Semi-Supervised Setting}
\label{method:semi}
As we will show in Section~\ref{sec:results}, despite better control over the sparsity-accuracy trade-off, there is still a gap in task performance between our unsupervised approach and a model that uses full context. To bridge this gap and better manage the trade-off at minimal annotation cost, we experiment with a semi-supervised setting where we have annotated rationales for part of the training data.

For input example $x = (x_{1}, x_{2}, \ldots , x_{n})$ and a gold mask $\hat{m} = (\hat{m_{1}}, \hat{m_{2}}, \ldots , \hat{m_{n}})$ over sentences, we use the following semi-supervised objective:
\begin{multline}
L_{semi} = \mathbf{E}_{m \sim p_{\theta}(m|x)}[-\log{q(y|m \odot x)}] + \\
\gamma \sum_j -\hat{m_j}\log{p(m_j|x)}
\label{semi1}
\end{multline}
While we still sample from $p(m|x)$ and train end-to-end using reparameterization, the information loss over $p(m|x)$ is replaced with the supervised rationale loss.


\section{Experimental Setup}

\subsection{End Tasks}
We evaluate our \ourmodel~approach on five text classification tasks from the ERASER benchmark \cite{deyoung2019eraser} and the \beer~regression task~\cite{mcauley2012learning} used in~\citet{lei2016rationalizing}.
\begin{itemize}[label={},leftmargin=0pt]
    \item \textbf{ERASER}: The ERASER tasks we evaluate on include the Movies sentiment analysis task~\cite{pang2004sentimental}, the FEVER fact extraction and verification task~\cite{thorne2018fever}, the MultiRC~\cite{khashabi2018looking} and BoolQ~\cite{clark2019boolq} reading comprehension tasks, and the Evidence Inference classification task~\cite{lehman2019inferring} over scientific articles for results of medical interventions.
    \item \textbf{\beer} \cite{mcauley2012learning}: The \beer~regression task  for predicting 0-5 star ratings for multiple aspects like appearance, smell, and taste based on reviews.
\end{itemize}
All these datasets have \emph{sentence-level} rationale annotations for validation and test sets. We do not consider e-SNLI \citep{camburu2018snli} and CoS-E \citep{rajani2019explain} in ERASER as they have only 1-2 input sentences, rationales annotations at word level, and often require common sense/world knowledge. The ERASER tasks contain rationale annotations for the training set, which we only use for our semi-supervised experiments.
We closely follow dataset processing in the ERASER benchmark setup and \citet{bastings2019interpretable} (for \beer). Additionally, for BoolQ and Evidence Inference which contain longer documents, we use a sliding window to select a single document \emph{span} that has the maximum TF-IDF score against the question (further details in Appendix~\ref{appendix:datasetprocessing}).

\subsection{Setup}

\paragraph{Evaluation Metrics} We adopt the metrics proposed for the ERASER benchmark to evaluate both agreement with comprehensive human rationales as well as end task performance. To evaluate quality of rationales, we report the token-level Intersection-Over-Union F1 (IOU F1), which is a relaxed measure for comparing two sets of text spans. We also report
token-level F1 scores. For task accuracy, we report weighted F1 for classification tasks, and the mean square error for the \beer~regression task.

\paragraph{Implementation Details}
We use BERT-base with a maximum context-length of 512 to instantiate the combined explainer and end-task predictor.
Models are tuned on the development set using the rationale IOU F1. Appendix \ref{appendix:hyperparameters} contains details about hyperparameters.

\subsection{Baselines}

\begin{table*}[ht]
    \centering
    \small
    \begin{tabular}{l|ccc|ccc|ccc}
        \toprule
        \bf Approach  & \multicolumn{3}{c|}{\bf FEVER} & \multicolumn{3}{c|}{\bf MultiRC} &
        \multicolumn{3}{c}{\bf Movies}\\
        & Task & Token F1 & IOU & Task &  Token F1 & IOU & Task & Token F1 & IOU \\
        \midrule
        1. \fullcontext & 89.5 & 33.7 & 36.2 & 66.8 & 29.1 & 29.2 & 91.0 & 35.1 & 47.3 \\
        2. \goldcontext & 91.8 & - & - & 76.6 & - & - & 97.0 & - & - \\
        \midrule
        \multicolumn{10}{c}{Unsupervised} \\
        \midrule
        3. \nosparsity & 82.8  & 35.7 & 38.1 & 60.1 & 20.8 & 19.8  & 78.2 & 24.6 & 37.9\\
        4. \sparsenorm  & 83.1 & 40.9 & 44.0 & 59.7 & 19.9 &  20.4  & 78.6 & 23.5 & 34.7\\
        5. \sparsenormthreshold & 83.3 & 41.6 & 44.9 & 61.7 & 21.7 & 21.8 & 81.8 & 22.8 & 34.4\\
        6. \ourmodel~(Us) & \bf 84.7 & \bf 42.7 & \bf 45.5 & \bf 62.1 & \bf 24.9 & \bf 24.3 & \bf 84.0 & \bf 27.5 & \bf 39.6 \\
        \midrule
        \multicolumn{10}{c}{Supervised} \\
        \midrule
        7. Bert-To-Bert (Reported) & 87.7  & 81.2 & 83.5 & 62.4 & 39.9 & 40.9 &  82.4 & 14.5 & 7.5 \\
        8. Bert-To-Bert (Ours$^\epsilon$)  & 85.0  & 78.1 & \bf 81.7 & 63.3 & 41.2 & 41.6 &  86.0 & 16.2 & 15.7\\
        9. $25\%$ data (Us) & \bf 88.8 &  63.9 &  66.6 & \bf 66.4 & \bf 54.0  & \bf 54.4 & \bf 85.4 & \bf 28.2 & \bf 43.4\\
        \midrule
    \end{tabular}
    \label{tab:final_resuls1}
\end{table*}

\begin{table*}[ht]
    \centering
    \small
    \begin{tabular}{l|ccc|ccc|ccc}
        \midrule
        \bf  & \multicolumn{3}{c|}{\bf BoolQ} & 
        \multicolumn{3}{c}{\bf Evidence Inference} &  \multicolumn{3}{c}{\bf \beer} \\
        & Task & Token F1 & IOU & Task &  Token F1 & IOU & Task & Token F1 & IOU \\
        \midrule
        1. \fullcontext & 65.6 & 11.8 & 15.0 & 52.1 & 6.4 & 9.7 & .015 & 38.4 & 37.8  \\
        2. \goldcontext &  85.9 & - & - & 71.7 & - & - & - & - & - \\
        \midrule
        \multicolumn{10}{c}{Unsupervised} \\
        \midrule
        3. \nosparsity &  62.5 &  8.1 & 10.7 & 43.0 & 6.1 & 09.0 & .018 & 48.2 & 47.3\\
        4. \sparsenorm  & 62.5 & 8.5 & 12.8 & 38.9 & 3.4 & 6.3 & .017 & 28.6 & 35.5\\
        5. \sparsenormthreshold & 63.7 & 10.7 & 14.3 & 44.7 & 5.1 & 8.0  & .018 & 49.3 & 49.0 \\
        6. \ourmodel~(Us) & \bf 65.2 & \bf 12.8 & \bf 16.5 & \bf 46.3 & \bf 6.9 & \bf 10.0 & \bf .016 & \bf 53.1 & \bf 52.3 \\
        \midrule
        \multicolumn{10}{c}{Supervised} \\
        \midrule
        8. Bert-To-Bert (Reported) & 54.4 & 13.4 &  5.2 & 70.8 & 46.8 & 45.5  &  &  & \\
        7. Bert-To-Bert (Ours$^\epsilon$)  & 62.3 & 18.4 & 31.5 & \bf 70.8 & \bf 54.8 & \bf  53.9  &  & $^\dagger$ & \\
        9. $25\%$ data (Us) & \bf 63.4 & \bf 19.2 & \bf 32.3 &  46.7 & 10.8 & 13.3 & & & \\
        \bottomrule
    \end{tabular}
    \caption{Task, Rationale IOU F1 (threshold set to 0.1) and Token F1 for our hard-attention \ourmodel~approach and baselines on test sets, averaged over 5 random seeds. We report MSE for \beer, hence lower is better. Gold IOU and token F1 are 100.0. We use $25\%$ training data in our semi-supervised setting (Section \ref{method:semi}). Validation set results can be found in  Table~\ref{tab:dev_results} in the Appendix. $^\epsilon$ We could not reproduce  numbers for the Bert-to-Bert supervised method reported in \citet{deyoung2019eraser}. $^\dagger$ No rationale supervision available for \beer.} 
    \label{tab:final_resuls2}
\end{table*}

\label{baselines}
We first consider two bounding scenarios where no rationales are predicted. In the \textbf{Full Context (\fullcontext)} setting, the entire context is used to make predictions; this allows us to estimate the loss in performance as a result of interpretable hard attention models that only use $\pi\%$ of the input. In the \textbf{Gold Rationale (\goldcontext)} setting, we train a model to only use human rationale annotations during \emph{training and inference} to estimate an upper-bound on task and rationale performance metrics. 
We compare our \ourmodel~approach with the following  baselines. For fair comparison, all baselines are modified to use BERT-based representations. 

\paragraph{Norm Minimization (\sparsenorm)}
Existing approaches ~\cite{lei2016rationalizing,bastings2019interpretable} learn sparse masks over the inputs by minimizing the $L_0$ norm of the mask $m$ as follows:
\begin{equation}
    L_{SL0} = \mathbf{E}_{m \sim p(m|x)}[-\log{q(y|z)}] + \lambda ||m||
\label{lsr}
\end{equation}
\noindent Here, $\lambda$ is the weight on the norm.
\paragraph{Controlled Norm Minimization (\sparsenormthreshold)} For fair comparison against our approach for controlled sparsity, we modify Equation~\ref{lsr} to ensure that the norm of $m$ is not penalized when it drops below the threshold $\pi$. 
\begin{multline}
     L_{SL0-C} = \mathbf{E}_{m \sim p(m|x)}[-\log{q(y|z)}] + \\
     \lambda \max{( 0, ||m|| - \pi)}
\label{lsr_pi}    
\end{multline}
This modification has also been adopted in recent. work \cite{jain2020learning}. Explicit control over sparsity in the mask $m$ through the tunable prior probability $\pi$  naturally emerges from IB theory, as opposed to the modification adopted in norm-based regularization (Equation~\ref{lsr_pi}).

\paragraph{\nosparsity} This method only optimizes for the end-task performance without any sparsity-inducing loss term, to evaluate the effect of sparsity inducing objectives in \ourmodel, \sparsenorm, and \sparsenormthreshold.

\paragraph{Supervised Approach (Pipeline)} \citet{lehman2019inferring} learn an explainer and a task predictor independently in sequence using supervision for rationales and task labels, using the output of the explainer in the predictor during inference. We compare our semi-supervised model (Section~\ref{method:semi}) with this \emph{pipeline} approach.


\section{Results}
\label{sec:results}
\subsection{Quantitative Evaluation}
Table~\ref{tab:final_resuls2} compares our  \ourmodel~approach against baselines (Section~\ref{baselines}).
\ourmodel~outperforms norm-minimization approaches (rows 4-6) in both agreement with human rationales and task performance across all tasks. We perform particularly well on rationale extraction with 5 to 80\% relative improvements over the better performing norm-minimization variant \sparsenormthreshold. \ourmodel~also attains task performance within 0.5 to 10\% of the full-context model (row 1), despite using $<50\%$ of the input sentences. All unsupervised approaches still obtain a lower IOU F1 compared to the full context model for Movies and MultiRC, primarily due to their considerably lower precision on these benchmarks.

\begin{figure*}[t]
    \centering
    \includegraphics[scale=0.30]{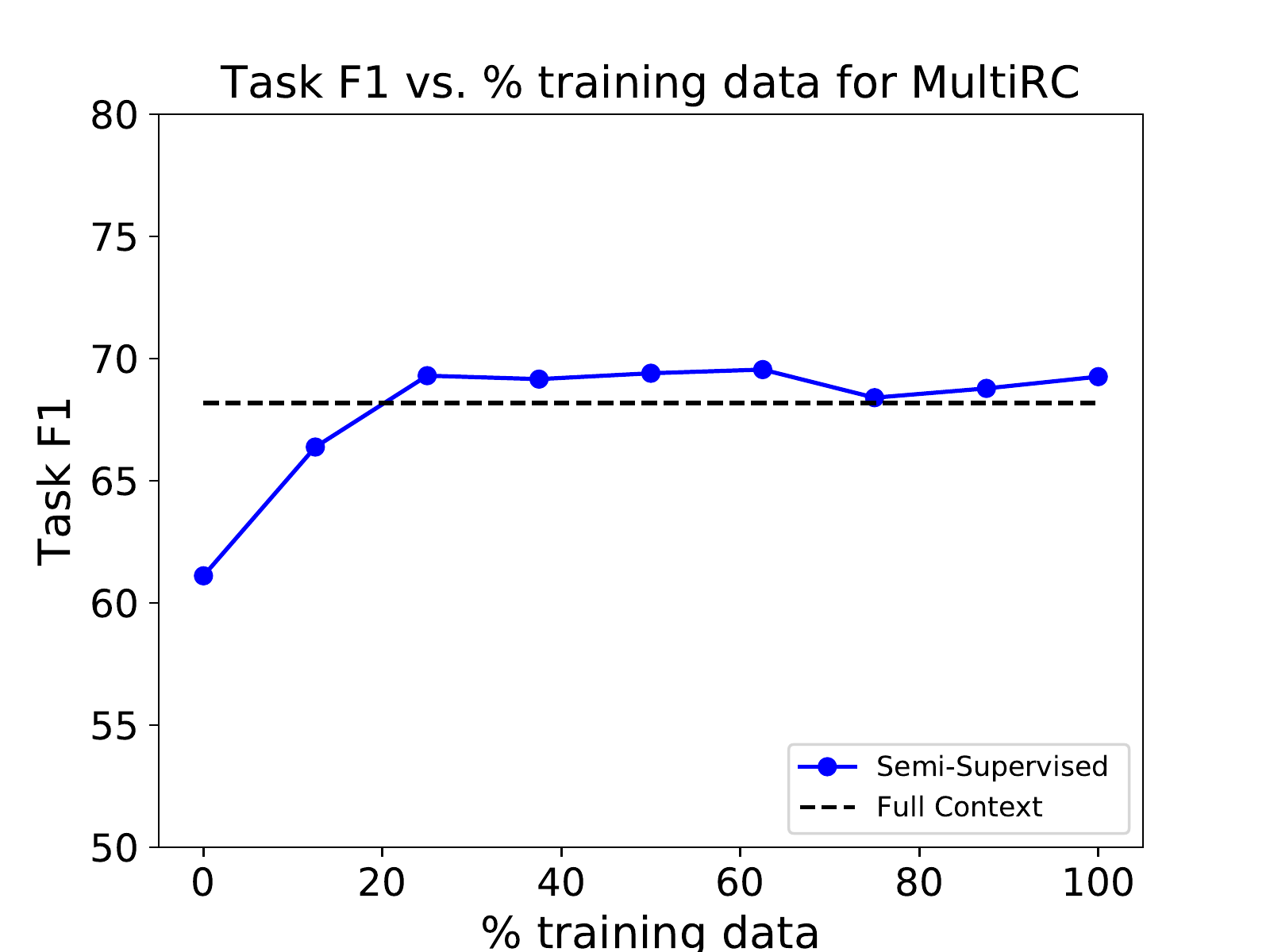}
    \includegraphics[scale=0.30]{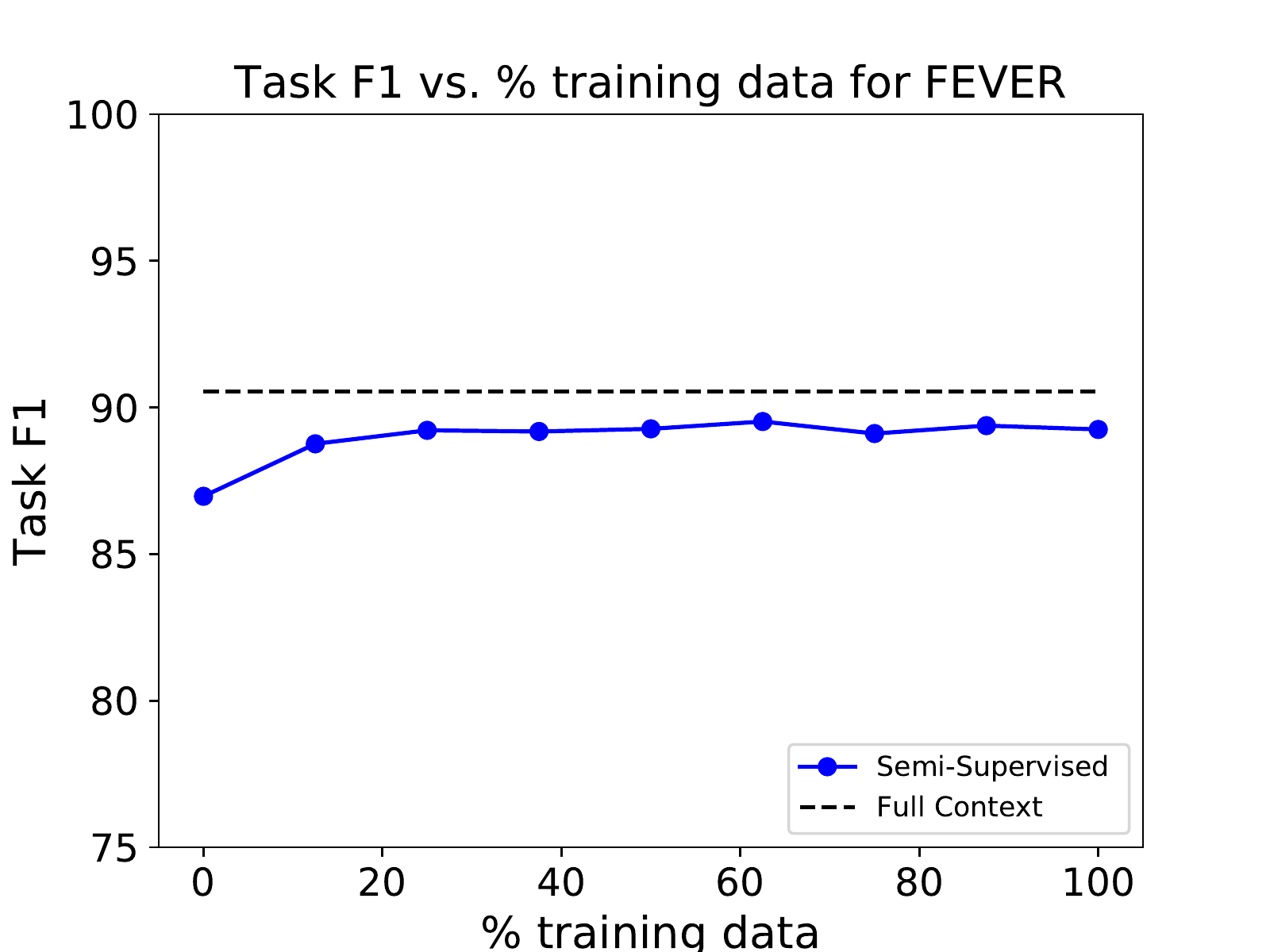}
    \includegraphics[scale=0.30]{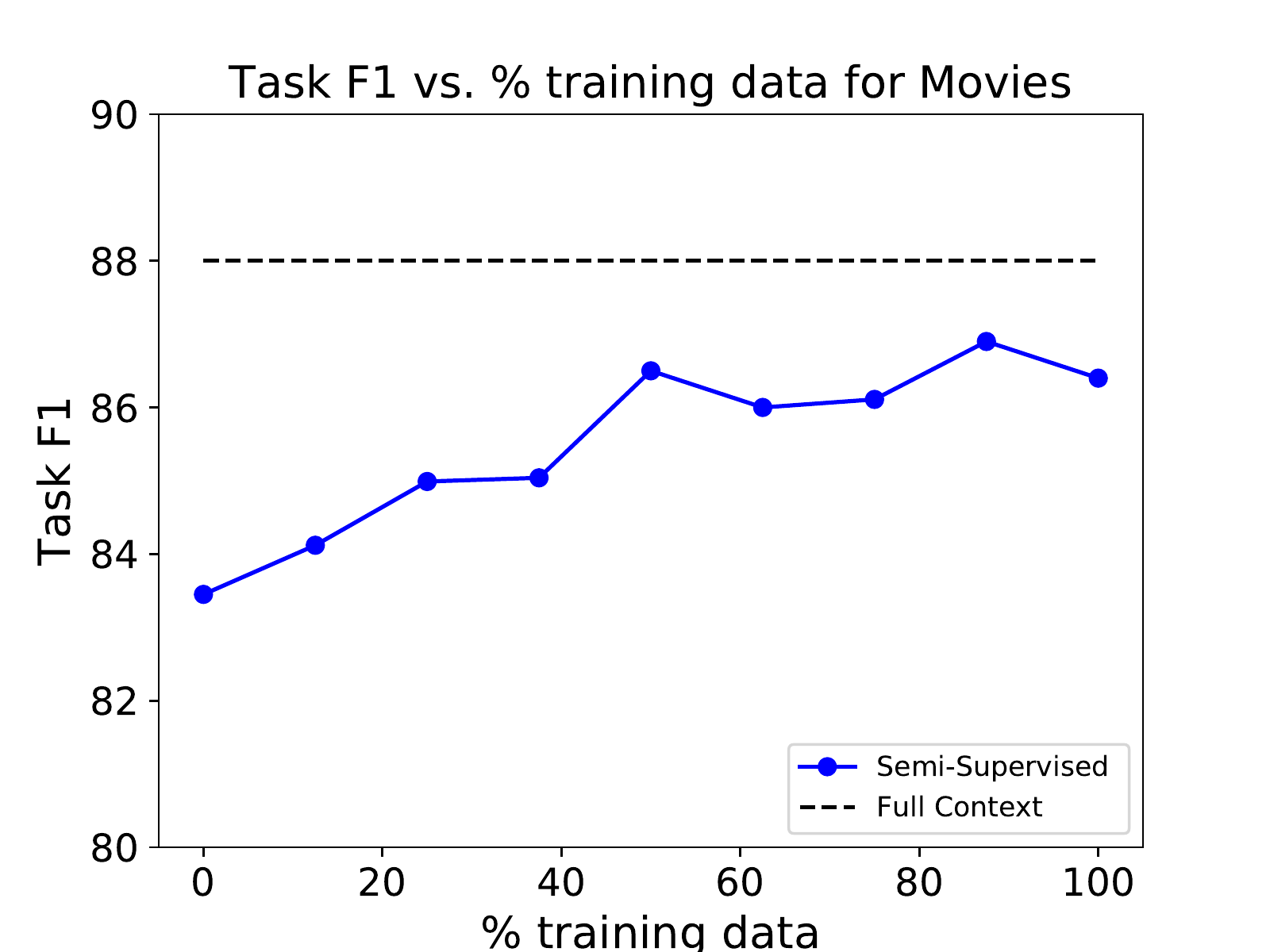}
    \caption{Semi-supervised experiments showing the task performance for varying proportions of rationale annotation supervision on the MultiRC, FEVER, and Movies datasets.}
\end{figure*}

\begin{figure}[t]
    \centering
    \includegraphics[scale=0.50]{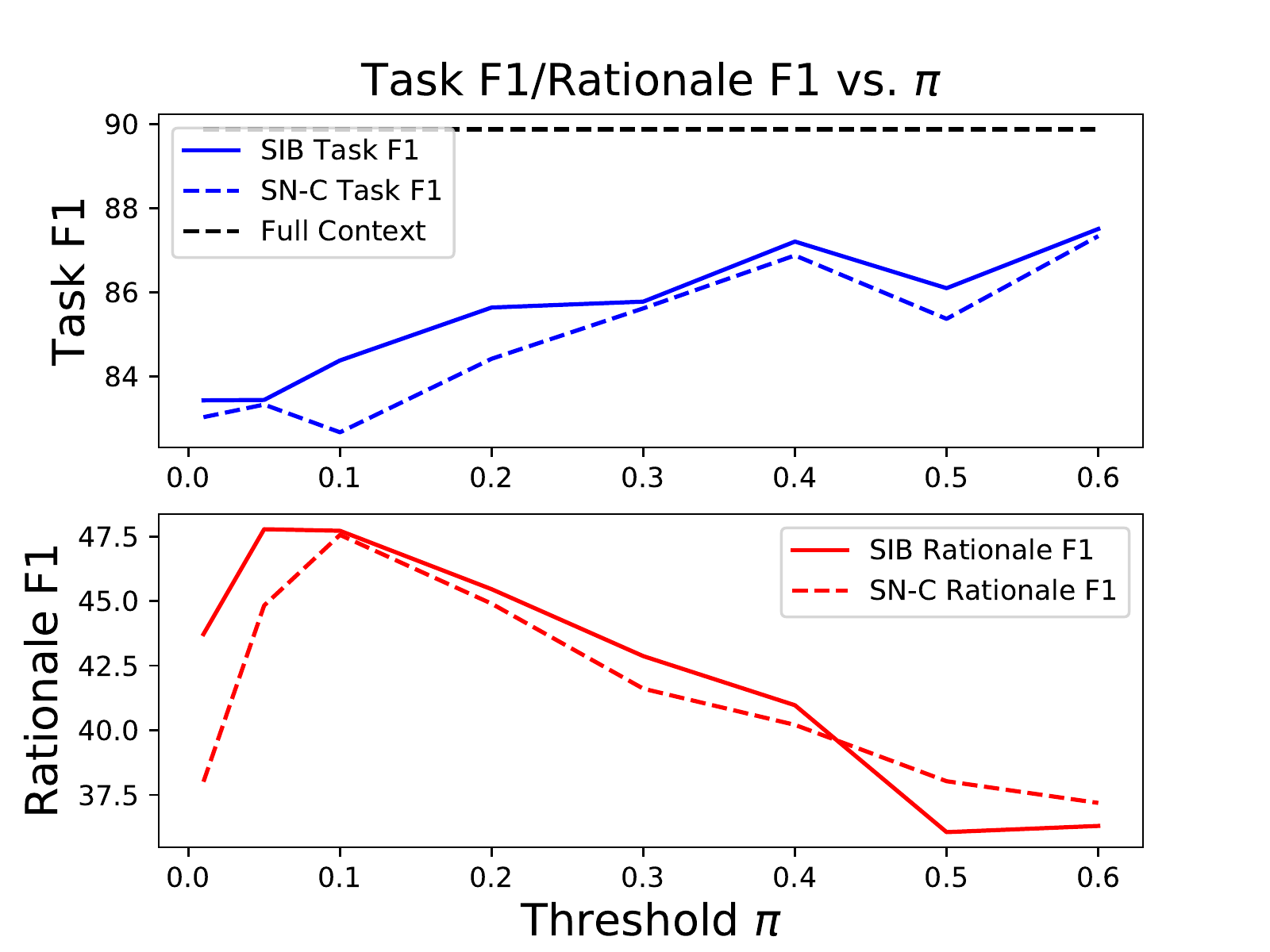}
    \caption{Effect of varying the sparsity hyperparameter $\pi$ to control the trade-off between compactness of rationales and accuracy for the FEVER dataset (right). SIB is \ourmodel~ and SN-C is \sparsenormthreshold.
    }
    \label{fig:semi}
\end{figure}

Our results also highlight the importance of explicit \emph{controlled sparsity} inducing terms as effective inductive biases for improved task performance and rationale agreement. Specifically, sparsity-inducing methods consistently outperform the \nosparsity-baseline (row 3). One way to interpret this result is that sparsity objectives add input-dimension regularization during training, which results in better generalization during inference. 
Moreover, \sparsenormthreshold, which adds the element of \emph{control} to norm-minimization,  performs considerably better than \sparsenorm. 
Finally, we see a positive correlation between task performance and agreement with human rationales. 
This is important since accurate models that also better emulate human rationalization likely engender more trust.


\paragraph{Semi-supervised Setting}
In order to close the performance gap with the full-context model, we also experiment with a setup where we minimize the task and the rationale prediction loss using rationale annotations available for a part of the training data (Section~\ref{method:semi}).
Figure~\ref{fig:semi}~(left, center) shows the effect of incorporating an increasing proportion of rationale annotation supervision for the FEVER and MultiRC datasets. Our semi-supervised model is even able to match the performance of the full-context models for both FEVER and MultiRC with only 25\% of rationale annotation supervision. Furthermore, Figure \ref{fig:semi} also shows that these gains can be achieved with relatively modest annotation costs since adding more rationale supervision to the training data seems to have diminishing returns. 

Table \ref{tab:final_resuls2} compares our interpretable model (row 9), which uses rationale supervision for 25\% of the training data, with the full-context model and the 
Pipeline approach (row 8).
On three (FEVER, MultiRC, and BoolQ) out of five datasets for which rationale supervision is available, our interpretable models match the task performance of the full-context models while recording large gains in IOU (17-30 F1 absolute). 
Our approach outperforms the pipeline-based approach in task performance (for FEVER, MultiRC, Movies, and BoolQ) and IOU (for MultiRC and Movies).
These gains may result from better exploration due to sampling and inference based on a fixed budget of $\pi\%$ sentences.
Our weakest results are on Evidence Inference where the TF-IDF preprocessing often fails to select relevant rationale spans and the pipeline approach uses SciBERT \citep{beltagy2019scibert}.\footnote{Only 51.8\% of the selected passages have gold rationales.}
Our overall results suggest that a small proportion of direct supervision can help build interpretable models without compromising task performance.

\subsection{Analysis}
\label{subsec:analysis}

\begin{table}[t]
    \centering
    \small
    \begin{tabular}{lccccc}
    \toprule
    Dataset & $\pi$ & \multicolumn{2}{c}{\sparsenormthreshold} & \multicolumn{2}{c}{\ourmodel} \\
    & & Mean & Var & Mean & Var \\
    \midrule
    FEVER & 0.20 & 0.17 & 0.94 & 0.21 & 1.24 \\
    MultiRC & 0.25 & 0.11 & 1.14 & 0.26 & 1.67\\
    Movies & 0.40 & 0.38 & 2.90 & 0.42 & 3.02\\
    BoolQ & 0.20 & 0.04 & 0.84 & 0.22 & 1.91\\
    Evidence & 0.20 & 0.10 & 1.17 & 0.20 & 1.61 \\
    \bottomrule
    \end{tabular}
    \caption{Average mask length (sparsity) attained by \ourmodel~ and the \sparsenormthreshold~baseline for a given prior $\pi$ for different tasks, averaged over 100 runs. Mean is reported as the average proportion of sentences to compare with expected sparsity ($\pi$) and variance is reported in the number of sentences.}
    \label{tab:threshold}
\end{table}

\begin{table*}[ht]
    \centering
    \small
    \begin{tabular}{p{15.5cm}}
    \toprule
    Examples from Error Analysis \\
    \midrule
    Prediction:Positive \\
    Ground Truth:Negative \\
    \textcolor{red}{The original Babe gets my vote as the best family film since the princess bride, and it's sequel has been getting 
    rave reviews} \textcolor{red}{from most internet critics, both Siskel and Ebert sighting it more than a month ago as one of the 
    year's finest films.} So, 
    naturally, when I entered the screening room that was to be showing the movie and  
    there was nary another viewer to be
    found, this notion left me  puzzled. \textcolor{red}{It is a rare thing for a children's movie to be praised this highly}
    $\dots$   Looking back, \textcolor{blue}{I should have taken the
     hint}  \textcolor{blue}{and left right when I entered the theater}. Believe me; I wanted to like Babe: Pig in the City. \textcolor{red}{The plot
     seemed interesting enough;} $\dots$  It is here that we meet an array of eccentric characters, the most memorable 
     being the family of chimps led by Steven Wright. \textcolor{blue}{Here is where the film took a wrong turn  $\dots$
     unfortunately, the} \textcolor{blue}{story wears thin as we are introduced to a new set of animals that} $\dots$ the main topic of
     discussion $\dots$ \textcolor{blue}{it just didn't feel right} \textcolor{blue}{and was more painful to watch} than it was funny or entertaining, and the same goes for the rest of the movie. \\
     \midrule
     Statement : Unforced labor is a reason for human trafficking.\\
Prediction: SUPPORTS \\
Ground Truth: REFUTES\\
DOC: \textcolor{blue}{Human trafficking is the trade of humans, most commonly for the purpose of \emph{forced} labour, sexual slavery, or comm-}\textcolor{blue}{-ercial 
sexual exploitation for the trafficker or others.} This may encompass providing a spouse in the context of forced marriage, or the 
extraction of organs or tissues, including for surrogacy and ova removal. Human trafficking can occur within a country or transnationally. 
coercion and because of their commercial exploitation $\dots$  In 2012, the I.L.O. estimated that
21 million victims are trapped in 
modern-day slavery $\dots$\\
\midrule
Statement: Atlanta metropolitan area is located in south Georgia.\\
Prediction: SUPPORTS\\
Ground Truth:REFUTES\\
DOC: \textcolor{purple}{Metro Atlanta , designated by the United States Office of Management and Budget as the Atlanta-Sandy Springs-}\\ \textcolor{purple}{Roswell, GA Metropolitan Statistical Area, is the most populous metro area in the US state of Georgia and the ninth-largest } \textcolor{purple}{metropolitan statistical area (MSA) in the United States}. Its economic, cultural and demographic center is 
Atlanta, and it had a 2015 estimated population of 5.7 million people according to the U.S. Census Bureau. The metro  area forms the core of a broader trading area, the Atlanta -- Athens-Clarke -- Sandy Springs Combined Statistical Area. \textcolor{blue}{The Combined Statistical Area spans up to 39 counties in north Georgia and had an estimated 2015 population of 6.3 million}  \textcolor{blue}{people}. Atlanta is considered an `` alpha world city ''. \textcolor{red}{It is the third largest metropolitan region in the Census Bureau's } \textcolor{red}{Southeast region behind Greater Washington and South Florida.}\\
\bottomrule
    \end{tabular}
    \caption{Misclassified examples from the Movies and FEVER datasets show: (a) limitations in considering more complex linguistic phenomena like sarcasm; (b) overreliance on shallow lexical matching---\emph{un}forced vs. forced; 
    (c) limited world knowledge---south Georgia, Southeast region, South Florida.
    \emph{Legend}:  \textcolor{red}{Model evidence}, \textcolor{blue}{Gold evidence}, \textcolor{purple}{Model and Gold Evidence}}
    \label{fig:qualitative_ex} 
\end{table*}

\paragraph{Accurate Sparsity Control}
\label{sparsity_analysis}
Table~\ref{tab:threshold} compares average sparsity rates in rationales extracted by \ourmodel~with those extracted by norm-minimization methods. We measure the sparsity achieved by the explainer during inference by computing the average number of \emph{one} entries in the input mask $m$ over sentences (the hamming weight) for 100 runs. 
\ourmodel~consistently achieves the sparsity level $\pi$ used in the prior while the norm-minimization approach (\sparsenormthreshold) converges to a lower average sparsity for the mask.  


\paragraph{Sparsity-Accuracy Trade-off}
Figure \ref{fig:semi}~(right) shows the variation in task and rationale agreement performance as a function of the sparsity rate $\pi$ for \ourmodel~and \sparsenormthreshold~on the FEVER dataset. Both methods extract longer rationales with increasing $\pi$ that results in a decrease in agreement with sparse human rationales, while accuracy improves. However, \ourmodel~consistently outperforms \sparsenormthreshold~in task performance.

In summary, our analysis indicates that unlike norm-minimization methods, our IB objective is able to consistently extract rationales with the specified sparsity rates, and achieves a better trade-off with accuracy. We hypothesize that optimizing the KL-divergence of the posterior $p(m|x)$ may be able to model input salience better than an implicit regularization (through $||m||_0$). The sparse prior term can learn $p(m|x)$ adaptive to different examples, while $||m||$ encourages uniform  sparsity across examples.\footnote{Unlike the norm $||m||_0$, the derivative of KL-divergence term is proportional to $\log{p(m|x)}$} This can be seen explicitly in Table  \ref{tab:threshold}, where the variance in sampled mask across examples is higher for our objective.

\paragraph{Model Agnostic Behavior}
Our approach is agnostic to choice of model architecture and word vs. sentence level rationales. We experimented with the word-level model in \cite{deyoung2019eraser}, where masks are learned over words instead of sentences. 
More details of the model architecture can be found in  \cite{deyoung2019eraser}.
The results for which are shown in Table \ref{tab:model_agnostic}

\begin{table}[ht!]
    \centering
    \small
    \begin{tabular}{lcccc}
        \toprule
        \bf Approach  & \multicolumn{2}{c}{\bf Movies} & \multicolumn{2}{c}{\bf MultiRC} \\
        & Task & IOU & Task & IOU \\
        \midrule
        \sparsenormthreshold & 91.96  & 48.9 & 64.25 &  25.7 \\ 
        \ourmodel~(Us) & \bf 93.46 & \bf 52.1 & \bf 65.63 & \bf  27.0 \\
        \bottomrule
    \end{tabular}
    \caption{Task and IOU F1 for our \ourmodel~approach and best performing baseline on word-level rationales and BERT+LSTM model.} 
    \label{tab:model_agnostic}
\end{table}

\paragraph{Error Analysis}
A qualitative analysis of the rationales extracted by the \ourmodel~approach indicates that such methods struggle when the context offers spurious---or in some cases even genuine but limited---evidence for both output labels (Figure \ref{fig:qualitative_ex}). For instance, the model makes an incorrect positive prediction for the first example from the Movies sentiment dataset based on sentences that praise the \emph{prequel} of the movie or acknowledge some critical acclaim.
We also observed incorrect predictions based on shallow lexical matching (likely equating \emph{forced} and \emph{unforced} in the second example)
and world knowledge (likely equating south \emph{Georgia}, southeastern \emph{United States}, and South \emph{Florida} in the third). 
Overall, there is scope for improvement through better incorporation of exact lexical match, coreference propagation, and representation of pragmatics in our sentence representations.




\section{Related Work}

\paragraph{Extractive Rationalization}
Methods that condition predictions on their explanations are more trustworthy than post-hoc explanation techniques \cite{ribeiro2016should, krause2017workflow, alvarez2017causal} and analyses of self-attention \cite{serrano2019attention, jain2020learning}.
Extractive rationalization \cite{lei2016rationalizing} is one of the most well-studied of such methods and has received increased attention with the recently released ERASER benchmark \cite{deyoung2019eraser}.
\citet{chang2019game} and  \citet{yu2019rethinking, chang2019game} have complementary work on class-wise explanation extraction.
\citet{bastings2019interpretable} employ a reparameterizable version of the 
bi-modal 
beta distribution (instead of Bernoulli) for the binary mask. 
While our method has focused on unsupervised settings due to the considerable cost of obtaining reliable rationale annotations, recent work \cite{lehman2019inferring} has also attempted to use direct supervision from rationale annotations for critical medical domain tasks.
Finally, \citet{latcinnik2020explaining} and \citet{rajani2019explain} focus on generating explanations (rather than extracting them from the input). The extractive paradigm can be unfavourable for certain ERASER tasks like commonsense question answering, where the given input provides limited context for the task.

\paragraph{Information Bottleneck}
Information Bottleneck (IB)~\cite{tishby99information} has recently been adapted in a number of downstream applications like parsing \cite{li2019specializing}, extractive summarization \cite{west2019bottlesum}, and image classification \cite{alemi2016deep, zhmoginov2019information}.
\citet{alemi2016deep} and \citet{li2019specializing} use IB for optimal compression of hidden representations of images and words respectively. We are interested in compressing the number of cognitive units (like sentences) 
to ensure interpretability of the bottleneck representation, similar to  \citet{west2019bottlesum}.
However, while \citet{west2019bottlesum} use brute-force search to optimize IB for summarization, we directly optimize a parametric variational bound on IB for rationales.
IB has also been previously used for interpretability---\citet{zhmoginov2019information} use a VAE to estimate the prior distribution over $z$ for image classification. 
\citet{bang2019explaining} use IB for post-hoc explanation of sentiment classification. They do not enforce a sparse prior, and as a result, cannot guarantee that the rationale is strictly smaller than the input. Controlling sparsity to manage the accuracy-conciseness trade-off is also not possible in their model.

\section{Conclusion}
We introduce a novel sparsity-inducing objective derived from the Information Bottleneck principle to extract rationales of desired conciseness. Our approach outperforms existing norm-minimization techniques in task performance and agreement with human rationales for tasks in the ERASER benchmark.
Our objective obtains a better trade off of accuracy vs. sparsity. 
We are also able to close the gap with models that use the full input with $<25\%$ rationale annotations for a majority of the tasks.
In future work, we would like to 
 apply our approach on document-level and multi-document NLU tasks.

\section*{Acknowledgments}
This research was supported by ONR N00014-18-1-2826, DARPA N66001-19-2-403, ARO
W911NF-16-1-0121 and NSF IIS-1252835,
IIS-1562364, an Allen Distinguished Investigator Award, and the Sloan Fellowship. 
We thank Prof. Sreeram Kannan, Andrey Zhmoginov,
and the UW NLP group for helpful conversations
and comments on the work.

\bibliography{emnlp-ijcnlp-2019}
\bibliographystyle{acl_natbib}

\clearpage
\appendix

\section{Information Bottleneck Theory}
We first present an overview of the variational bound on IB introduced by \cite{alemi2016deep} and then derive a modified version amenable to interpretability.

\subsection{Variational Information Bottleneck (\citet{alemi2016deep})}
\label{ib_alemi}
The objective is to parameterize the information bottleneck objective $L_{IB} = I(X, Z) - \beta I(Z, Y)$ using neural models and use SGD to optimize. Consider the joint distribution: $p(X, Y, Z) = p(Z|X, Y )p(Y |X)p(X) = p(Z|X)p(Y |X)p(X)$ under the Markov chain $Y \leftrightarrow X \leftrightarrow Z$. As mutual information is hard to compute, the following bounds are derived on both MI terms:\\

First Term:\\
$$I(Z, X) := \mathbb{E}_x\left[\mathop{\mathbb{E}}_{z \sim p_{\theta}(z|x)}\left[\log{\frac{p_{\theta}(z|x)}{p(z)}}\right]\right]$$
where, 
$$p(z) := \int dx p_{\theta}(z|x)p(x)$$
This marginal is intractable. Let $r(z)$ be a variational approximation to this marginal. Since $\text{KL}[p(z), r(z)] \geq 0$, 
\begin{align*}
I(Z,X) \leq \mathbb{E}_x\left[\mathop{\mathbb{E}}_{z \sim p_{\theta}(z|x)}\left[\log{\frac{p_{\theta}(z|x)}{r(z)}}\right]\right]
\end{align*}
If $p_{\theta}(z|x)$ and $r(z)$ are of a form that KL divergence can be analytically computed, we get:
$$I(Z,X) \leq \mathbb{E}_x\left[\text{KL}[p_{\theta}(z|x), r(z)\right]$$
Typically, the distributions $p_\theta(z|x)$ and $r(z)$ are instantiated as multivariate Normal distributions to analytically compute the KL-divergence term.
$$r(z) = \mathcal{N}(z|0, I),  \;\;  p(z|x) = \mathcal{N}(z|{\mu}(x), {\Sigma}(x));$$
where $\mu$ is a neural network which outputs the K-dimensional mean of z and $\Sigma$ outputs the $K \times K$ covariance matrix $\Sigma$. This also allows us to reparameterize samples drawn from $p_\theta(z|x)$.\\

Second Term:\\
$$I(Z, Y) : = \mathop{\mathbb{E}}_{y,z \sim p_{\theta}}\left[\log{\frac{p(y|z)}{p(y)}}\right]$$
where, 
$$p(y|z) := \int dx \frac{p(y|x)p(z|x)p(x)}{p(z)}$$
Again, as this is intractable, $q_{\phi}(y|z)$ is used as a variational approximation to $p(y|z)$ and is instantiated as a transformer model with its own set of parameters $\phi$. As Kullback Leibler divergence is always positive:
$$\text{KL}[p(y|z), q_\phi(y|z)] \geq 0 \rightarrow$$
$$I(Z, Y) \geq \mathop{\mathbb{E}}_{y,z \sim p_{\theta}}\left[\log{\frac{q_{\phi}(y|z)}{p(y)}}\right]$$ 
The term $p(y)$ can be dropped as it is constant with respect to parameters $\phi$. Thus, we minimize  $\mathbb{E}_{y,z \sim p_{\theta}}[-\log{q_{\phi}(y|z)}]$
Thus the IB objective is bounded by the loss function:
$$L_{vib} \geq \mathbb{E}_{y,z \sim p_{\theta}}[-\log{q_{\phi}(y|z)}] + \beta \text{KL}[p_{\theta}(z|x), r(z)]$$

\begin{table*}[ht]
    \centering
    \begin{tabular}{ccccccc}
    \toprule
        \bf Hyperparameter & \bf Movie & \bf FEVER & \bf MultiRC & \bf BoolQ & \bf Evidence Inference & \bf BEER\\
    \midrule
        NS & 36 & 10 & 15 & 25 & 20 & 10 \\
        $\pi$ (Sparsity threshold (\%)) & .40 & .20 & .25 & .20 & .20 & .20 \\ 
        $\gamma$ (weight on SR)  & 0.5	& 0.05 & 	1.00E-04 &	0.01	& 0.001	& 0.01 \\ 
    \bottomrule
    \end{tabular}
    \caption{Hyperparameters used to report results.}
    \label{tab:hyper}
\end{table*}

\begin{table*}[ht]
    \centering
    \begin{tabular}{c|cc|cc|cc|cc|cc}
        \toprule
        \bf Approach  & \multicolumn{2}{c|}{\bf FEVER} & \multicolumn{2}{c|}{\bf MultiRC} &
        \multicolumn{2}{c|}{\bf Movies} &
        \multicolumn{2}{c|}{\bf BoolQ} & \multicolumn{2}{c}{\bf Evidence}
        \\
        & Task & IOU & Task & IOU & Task & IOU & Task & IOU & Task & IOU \\
        \midrule
        \fullcontext & 90.54 & - & 68.18 & - & 88.0 & - & 63.16 & - & 47.51 & - \\
        \goldcontext & 92.52 & - & 78.20 & - & 1.0  & - & 71.65 & - & 85.39 & - \\
        \nosparsity & 83.01	& 35.50 & 59.17 & 22.42 & 81.46 & 20.63 & 61.82 & 10.39 & 47.51 &	9.87 \\
        \sparsenorm & 84.30 & 45.44 & 58.40 & 20.41 & 79.35 & 19.23 & 59.04 & 12.40 & 44.52 & 9.4 \\
        \sparsenormthreshold & 84.42  & 44.90 & 60.77 & 23.25 & 82.43 & 18.91 & 62.24 & 09.72 & 48.97 & 09.40 \\
        \ourmodel & \bf 85.64 & \bf 45.46 & \bf 61.11 & \bf 25.55 & \bf 86.50 & \bf 22.33 & \bf 63.07 & \bf 16.63 & \bf 49.09 & \bf 11.09 \\
        \bottomrule
    \end{tabular}
    \caption{Final results of our unsupervised models
    on ERASER Dev Set}
    \label{tab:dev_results}
\end{table*}

\subsection{Deriving the Sparse Prior Objective}
\label{iib_spike_spike}

The latent space learned in Appendix \ref{ib_alemi} is not easy to interpret.
Instead we consider a masked representation of the form $z = m \odot x$, where $m_j \in \{0,1\}$ is a binary mask sampled from a distribution $p_\theta(m_j|x) = \text{Bernoulli}(\theta_j(x))$. This is an adaptive masking strategy, defined by data-driven relevance estimators $\theta_j(x)$. The distributions over $x$ and $m$ induce a distribution on $z = m \odot x$ defined by the conditionals
\[
p_\theta(z_j|x) = (1-\theta_j(x))\delta(z_j) + \theta_j(x)\delta(z_j - x_j).
\]

Our prior, based on human annotations, is that rationale needed for a prediction is sparse; we encode this prior as a distribution over masks $r(m_j) = \text{Bernoulli}(\pi)$. The prior also induces a distribution on $z = m \odot x$ given by
\[
r(z_j|x) = (1-\pi)\delta(z_j) + \pi\delta(z_j - x_j).
\]

We want to enforce a constraint $p_\theta(z_j)~=~r(z_j)$; i.e. that the marginal distribution $p_\theta(z_j) = \int p_\theta(z_j|x)p(x)\,dx$ matches our prior $r(z_j)$. This is difficult to do directly, but as in Appendix \ref{ib_alemi}, we can construct an upper bound the mutual information between $x$ and $z$:
\[
I(Z,X) \leq \mathop{\mathbb{E}}_{x \sim p}\left[\text{KL}[p_\theta(z|x),r(z)]\right].
\]
The inequality is tight if $r(z) = p_\theta(z)$. By optimizing to minimize mutual information $I(Z,X)$, we will implicitly learn parameters $\theta$ that approximate the desired constraint on the marginal.

In contrast to \citet{alemi2016deep}, our prior $r(z)$ has no parameters; rather than using an expressive model $r(z)$ to approximate the $p_\theta(z)$, we instead use the fixed prior $r(z)$ to force the learned conditionals $p_\theta(z|x)$ to assume a form such that the marginal $p_\theta(z)$ approximately matches the marginal of the prior, $\pi$. Average mask sparsity values in Table \ref{tab:threshold} corroborate this.

By a limiting argument, we can compute the divergence between $p_\theta(z|x)$ and $r(z)$:
\begin{align*}
\text{KL}&(p_\theta(z_j|x),r(z_j)) \\
&= (1 - \theta_j(x))\int \delta(z_j)\log \frac{p_\theta(z_j|x)}{r(z_j)}\,dz_j\\
&+ \theta_j(x)\int\delta(z_j - x_j)\log \frac{p_\theta(z_j|x)}{r(z_j)}\,dz_j\\
&= (1 - \theta_j(x))\log\frac{1-\theta_j(x)}{1-\pi} + \theta_j(x)\log\frac{\theta_j(x)}{\pi p(x)}\\
&= \text{KL}(p_\theta(m_j|x),r(m_j)) - \theta_j(x)\log p(x).
\end{align*}
The term $\text{KL}[p_\theta(m_j|x),r(m_j)]$ is a divergence between two Bernoulli distributions and has a simple closed form. If $\theta_j(x)$ and $\log p(x)$ are uncorrelated then
\[
\mathop{\mathbb{E}}_{x \sim q} \left[-\theta_j(x)\log p(x)\right] = \pi H(X).
\]
The term $\pi H(X)$ is constant with respect to the parameters $\theta$ and can be dropped.

We use the same, standard cross-entropy bound discussed in Appendix \ref{ib_alemi} to estimate $I(Z,Y)$, leading us to our variational bound on IB with interpretability constraints
\begin{align*}
L_{IVIB} &= \mathbf{E}_{m \sim p(m|x)}[-\log{q(y|m \odot x)}] \\ 
&+ \beta \sum_j KL[p_\theta(m_j|x)||r(m_j)].\\
\end{align*}


\section{Experimental Details}

\subsection{Data Processing}
\label{appendix:datasetprocessing}
The train, test and validation splits are the same as used in the ERASER benchmark \cite{deyoung2019eraser} and for the Beer Advocate dataset \cite{bastings2019interpretable}. In order to batch operations, we process the data so that each example has at most NS sentences. NS is fixed based on the average number of sentences in the development set of the respective task (see Table \ref{tab:hyper}).
Some dataset specific processing details are highlighted below: 
\begin{itemize}[label={},leftmargin=0pt]
    \item \textbf{FEVER}: ERASER adapts the original fact verification task as a binary classification of whether the given evidence supports or refutes a given claim.
    \item \textbf{MultiRC}:  The reading comprehension task with multiple correct answers is modified into a binary classification task for ERASER, where each (rationale, question, answer) triplet has a true/false label.
    \item \textbf{BoolQ}: A Boolean (yes/no) question answering dataset  over Wikipedia articles. Since most documents are considerably longer than BERT's maximum context window length of 512 tokens (3.3K tokens on average), we use a sliding window to select a single document \emph{span} that has the maximum TF-IDF score against the question.
     \item \textbf{Evidence Inference}: A three-way classification task over full-text scientific articles for inferring whether a given medical intervention is reported to either \emph{significantly increase}, \emph{significantly decrease}, or have \emph{no significant effect} on a specified outcome compared to a comparator of interest. We again apply the TF-IDF heuristic as the average number of tokens is a document is 4.6K.
    \item \textbf{BEER}: The Beer Advocate regression task  for predicting 0-5 star ratings for multiple aspects like appearance, smell, and taste based on reviews.  We report on the appearance aspect.
\end{itemize}

\subsection{Modeling}
\label{appendix:modeling}
For question answering tasks in ERASER. $s$ and $x$ are encoded together in the sequence $s \texttt{[SEP]} x$  while assuming that $s$ is fully unmasked i.e. $p_{\theta}(m_s|x)=1$. Once again, the sequence $s \texttt{[SEP]} m \odot x $ is used if query $s$ is available, i.e., we assume no masking over $s$ as it is assumed to be essential to predict $y$.
\paragraph{Semi-supervised:} Whenever train loss is not available, only task loss is used. Evaluation is still done based on $\pi$\% sentences, to fairly compare with unsupervised models.

\subsection{Hyperparameters}
\label{appendix:hyperparameters}
We use a sequence length of 512, batch size of 16 \footnote{We used 2 GeForce GTX TITAN X GPUs and Cuda 10.1} and Adam optimizer with a learning rate of 5e-5. We do not use warm-up or weight decay. We run all model for 20 epochs and set patience to 10 (over iterations). 
Hyper-parameter tuning is done on the validation set for the rationale  performance metric (IOU F1\footnote{Calculated as per the definition in \url{https://github.com/jayded/eraserbenchmark/blob/master/rationale_benchmark/metrics.py} for threshold 0.1}) on the development sets for ERASER tasks and on the test set for BEER (only test set contains rationale annotations). We tune the value of $\pi \in \{0.05, 0.1, 0.15, ... 0.50 \}$. We found that \ourmodel~approach is not as sensitive to the parameter $\beta$ and fix it to 1 to simplify experimental design.
For baselines, we tune the values of the Lagrangian multipliers, $\lambda \in  \{1e\mhyphen4, 5e\mhyphen4, 1e\mhyphen3, \dots , 1\}$ as norm-based techniques are more sensitive to $\lambda$. The value of the $\gamma$ hyperparameter in the semi-supervised setup was set to 1.0 to simplify design.
Instead of explicitely tuning or annealing the Gumbel softmax parameter, we fix it to 0.7 across all our experiments (including baselines). Hyperparameters for each dataset used for the final results are presented in Table \ref{tab:hyper}.\footnote{We observed some variation ($<$0.50 F1) in results across across GPUs, well within the difference observed between \ourmodel~and baselines.}

\section{Analysis}

\begin{table*}[h]
    \centering
    \begin{tabular}{ccccccc}
    \toprule
    \bf Approach & \multicolumn{3}{c}{\bf Movies} & \multicolumn{3}{c}{\bf Fever}\\
    & Task F1 & IOU F1 & Sparsity & Task F1 & IOU F1 & Sparsity \\ 
    \midrule 
        \sparsenormthreshold  with learned $\pi$ & 89.86 & 24.18 & 0.99 & 89.0 & 36.2 & 0.98 \\
        \ourmodel & 91.0 &	24.18 &	0.98 &	88.50 &	36.2 & 	0.96 \\
        \ourmodel with learned $\pi$ & 86.97 &	25.63 &	0.45 &	85.64 &	45.71 &	0.14 \\
    \bottomrule
    \end{tabular}
    \caption{Evaluation of learnable $\pi$. Results on Dev set}
    \label{tab:leaarnable_pi}
\end{table*}

\begin{table}[ht]
    \centering
    \footnotesize
    \begin{tabular}{cccccc}
    \toprule
        \bf Distribution/Approach & \multicolumn{2}{c}{\bf Movies} & \multicolumn{2}{c}{\bf Fever}\\
        & Task & IOU & Task & IOU \\
    \midrule
        Bernoulli (\sparsenormthreshold) & 79.4 & 18.3 &	83.3 & 44.9 \\
        Bernoulli  \ourmodel & \bf 81.5 & \bf 21.8 & \bf 84.7 & \bf 45.5 \\
        Kuma \sparsenormthreshold & 81.8 & 21.0 & 84.9 & 43.0\\
        Kuma \ourmodel & \bf 83.4 & \bf 21.5 & \bf 85.6 & \bf 45.5 \\
    \bottomrule
    \end{tabular}
    \caption{Results on the Kumaraswamy distribution from \cite{bastings2019interpretable} on Dev set}
    \label{tab:kuma}
\end{table}

\paragraph{Learning the Value of $\pi$}
Instead of tuning the value of $\pi$, we can alternately learn an appropriate value by allowing $\pi$ to be a learnable parameter in our implementation. In our experiments (see Table \ref{tab:leaarnable_pi}, we found that that the norm-minimization completely degenerates and learns a very high value of $\pi$, as the norm-loss in Equation \ref{lsr_pi} (Section \ref{baselines}) can still be minimized if both $||m||$ and $\pi$ are driven close to $1.0$. In our case, since $pi$ is now a learnable parameter, we have to minimize the following objective. 
\begin{multline}
L_{IVIB} = \mathbf{E}_{m \sim p_{\theta}(m|x)}[-\log{q_{\phi}(y|m \odot x)}] + \\
\beta \sum_j KL[p_{\theta}(m_j|x)||r(m_j)] + \pi H(x)
\label{ivib_full}
\end{multline}
The caveat here is that it requires another hyperparameter, namely the constant $H(x) = \lambda$ \footnote{We could alternately estimate this using a VAE, as done in \citet{zhmoginov2019information}}. This is not unlike \sparsenorm or \sparsenormthreshold where sparsity is controlled through the hyperparameter $\lambda$. In Table \ref{tab:leaarnable_pi}, we compare the \ourmodel objective with Equation~\ref{ivib_full} for Movies and FEVER. We find that optimizing Equation~\ref{ivib_full} actually allows us to control the trade-off because of the presence of the term $\pi H(x)$ that enforces a smaller value for $\pi$. The learned value of $\pi$ is close the tuned value in Table \ref{tab:hyper}, thus we choose to report our main results across all models on tuned $\pi$. 

\paragraph{A More Expressive Distribution}
\label{kuma}
\citet{bastings2019interpretable} compare against the best-known previous work on norm regularization \citet{lei2016rationalizing} by exploring the bi-modal Kumaraswamy distribution \cite{fletcher1996estimation} to replace the Bernoulli distribution. This more expressive distribution may be able to complement our approach, as KL-divergence for it can be analytically computed \cite{nalisnick2017stick} (Appendix \ref{kuma}).
The KL divergence between the Kumaraswamy and Beta distribution can be analytically computed, as done in this work \cite{nalisnick2017stick}. In Table \ref{tab:kuma}, we show results on Movies and FEVER datasets for this distribution, comparing \ourmodel against the \sparsenormthreshold baseline. We find that the superior performance of the KL-divergence loss term persists.



\end{document}